\documentclass[letterpaper]{article} 
\usepackage{aaai25}  
\usepackage{times}  
\usepackage{helvet}  
\usepackage{courier}  
\usepackage[hyphens]{url}  
\usepackage{graphicx} 
\usepackage{natbib}  
\usepackage{caption} 
\usepackage{algorithm}
\usepackage{algorithmic}

\usepackage{adjustbox}
\usepackage{amsmath}
\usepackage{adjustbox}
\usepackage{booktabs}
\usepackage{multirow}
\usepackage{graphicx}
\usepackage{makecell}
\usepackage{colortbl} 
\usepackage{xcolor} 
\usepackage{amsmath}
\usepackage{amsfonts}

\usepackage{newfloat}
\usepackage{listings}
\DeclareCaptionStyle{ruled}{labelfont=normalfont,labelsep=colon,strut=off} 
\floatstyle{ruled}
\newfloat{listing}{tb}{lst}{}
\floatname{listing}{Listing}
\title{PIP-MM: Pre-Integrating Prompt Information into Visual Encoding via Existing MLLM Structures}

\author {
    Tianxiang Wu\textsuperscript{\rm 1},
    Minxin Nie\textsuperscript{\rm 2},
    Ziqiang Cao\textsuperscript{\rm 3}
}
\affiliations {
    School of Computer Science \& Technology, Soochow University, SuZhou, China\\
    txwu@stu.suda.edu.cn, 
    123106010818@njust.edu.cn,
    zqcao@suda.edu.cn
}

\usepackage{bibentry}

\begin{document}

\maketitle

\begin{abstract}
The Multimodal Large Language Models (MLLMs) have activated the capabilities of Large Language Models (LLMs) in solving visual-language tasks by integrating visual information.
  The prevailing approach in existing MLLMs involves employing an image encoder to extract visual features, converting these features into visual tokens via an adapter, and then integrating them with the text prompt into the LLM.
  However, because the process of image encoding is prompt-agnostic, the extracted visual features only provide a coarse description of the image, impossible to focus on the requirements of the prompt.
  On one hand, these image features may sometimes overlook the objects specified in the prompt.
  On the other hand, the visual features contain a large amount of irrelevant information, which increases the memory burden and worsens the generation effectiveness.
  To address the aforementioned issues, we propose \textbf{PIP-MM}, a general framework that \textbf{P}re-\textbf{I}ntegrates \textbf{P}rompt information into the visual encoding process using existing modules of MLLMs.
  Specifically, We utilize the frozen LLM in the MLLM to vectorize the input prompt, which summarizes the requirements of the prompt.
  We then feed this prompt vector into our trained Multi-Layer Perceptron (MLP) to align it with the visual input criteria.
  This integration replaces the standard class embedding in the image encoder, enabling it to perceive and incorporate the prompt's directives into the visual encoding process.
  PIP-MM is parameter-efficient and can apply to various MLLMs.
  To validate its effectiveness, we undertook experiments on seven benchmarks, employing two different backbone MLLMs.
  Our method achieved an average performance improvement of 2.7\% over the baselines and demonstrated a 10\% higher win rate on artificially designed high-difficulty test sets.
  Moreover, our model maintains excellent generation results even when half of the visual tokens are reduced.
\end{abstract}

\begin{figure}[h]
\centering
\includegraphics[width=0.8\linewidth]{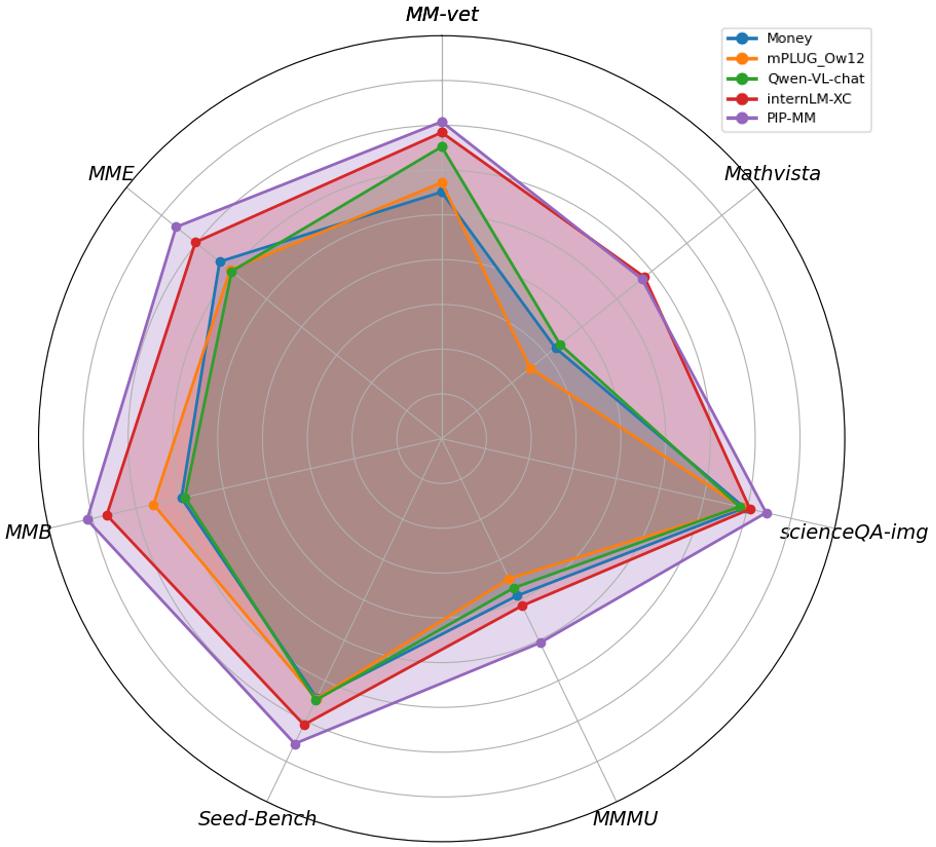}
\caption{
Compared to existing open-source SOTA models, PIP-MM performs on multiple visual-language task benchmarks.
}
\label{fig:radar}
\end{figure}

\section{Introduction}

\begin{figure*}[!t]
    \centering
    \includegraphics[scale=0.50]{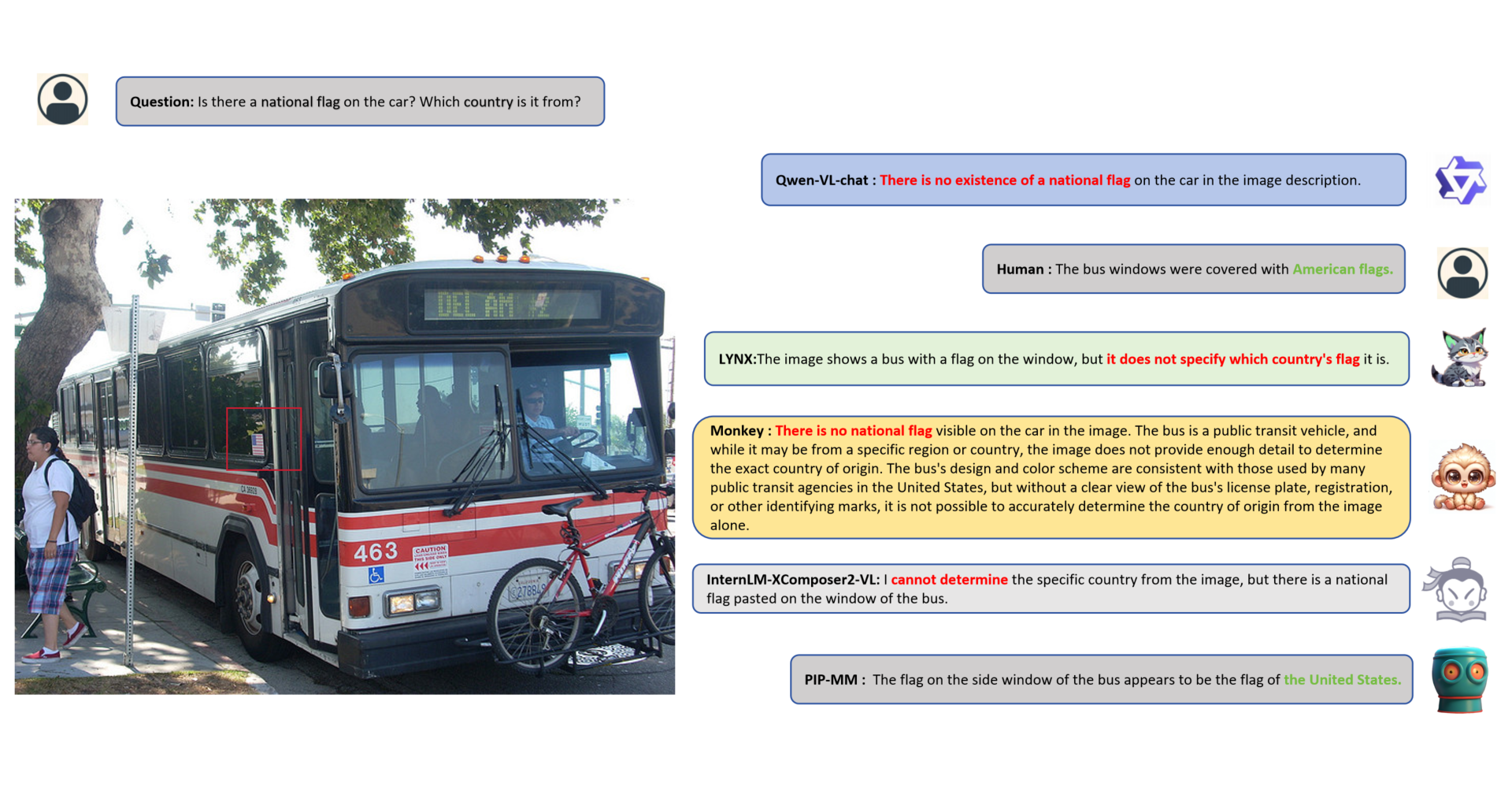}
    \caption{The performance of humans, some high-performing MLLMs, and PIP-MM under the confusion mode. The highlighted \textcolor{green}{green} portion represents the part where the image is correctly identified and the question is answered, while the highlighted \textcolor{red}{red} portion represents the parts where the model answers incorrectly or cannot be recognized.}
    \label{fig:intro}
\end{figure*}

In recent years, due to the outstanding generalization ability of LLMs in zero-shot tasks, researchers have been very active in studying LLMs \cite{llama,vicuna}.
At the same time, visual encoders have also been continuously developing in terms of image perception capabilities \cite{clip,vitsam,googlevit}.
Given the complementary nature between the two, the independent development of two single-modal domains has ultimately led to the emergence of the new field of MLLM \cite{2023MMsurvey}.
Existing MLLMs \cite{wang2023cogvlm,gpt4,bai2023qwen,internlmxcomposer} demonstrate excellent performance in tasks such as image description\cite{cc3m} and open visual question answering \cite{okvqa,docvqa,scienceqa}.

The current mainstream methods \cite{2023MMsurvey} involve using ViT to encode images, training an adapter to obtain visual tokens, and then inputting these tokens into an LLM to generate responses.
Finally, the visual tokens are concatenated with prompts to generate responses under the powerful LLM.
However, this prompt-unaware image encoding process suffers from two problems.
On the one hand, the extracted visual features may neglect the prompt-required information.
On the other hand, these visual features inevitably contain a significant amount of noise unrelated to the prompt, which not only affects the quality of text generation but also increases computational overhead.
These issues become particularly pronounced in scenarios where the image content is rich and the textual instructions mention secondary objects in the image (referred to as the \textbf{confusion mode} below).
As shown in Figure~\ref{fig:intro}, in a street traffic scene, the model's identification of the flag on the left side of the bus in the examples is unsatisfactory.
Although some works \cite{instructblip,bliva} have attempted to integrate prompts and visual features in the adapter section, they cannot handle the inherent problem of missing prompt-specified objects during image encoding.

Considering that the main issues arise in the visual encoding stage, we first focus on the class embedding (CLS) in ViT.
It plays a crucial role in capturing global information from images and leading downstream tasks like image classification \cite{vit2022survey}.
However, CLS is usually discarded after the ViT encoder in visual-language tasks.
Additionally, we notice that LLMs are capable of summarizing text into vectors \cite{chang2023llmsurvey}.
Based on the above analysis, we propose PIP-MM, a general framework that \textbf{P}re-\textbf{I}ntegrates \textbf{P}rompt into the image encoding process using existing modules of MLLMs.
The workflow of PIP-MM is as follows: first, the LLM reads and vectorizes the prompt.
Then we train a Multi-Layer Perceptron (MLP) as a text-to-image adapter to align the prompt vector with the visual input criteria.
Afterward, we replace CLS in ViT with the prompt vector, achieving an early fusion of text and image.
Finally, the prompt-aware visual features extracted by VIT are fed into the original modules of MLLMs to generate a response. 

We randomly sampled a large amount of data from datasets such as image caption and visual question answering (VQA) and trained PIP-MM using the classic two-stage training approach as described in \cite{bliva}.
Our training process is divided into two stages: In the first stage, we utilized a vast number of image-text pairs to train the MLP, ensuring that the text information extracted by the LLM is aligned with the image encoder.
In the second stage, we employed high-quality VQA datasets to fine-tune both the MLP and the adapter, thereby enhancing the MLLM's ability to follow instructions and process images.
Throughout the entire training process, the parameters we trained accounted for less than one percent of the entire model, making it an extremely efficient training method.
To substantiate the efficacy of PIP-MM, we conducted experiments across multiple benchmarks, including MM-Vet \cite{yu2023mmvet} and MME \cite{2023MMsurvey}.
Additionally, we performed tests on two distinct base models \cite{dong2024internlm2,zeng2024matters} to ensure the versatility of our approach.
Both automatic and manual evaluations confirm that our model outperforms all the other baselines.
Figure \ref{fig:radar} demonstrates the superior performance of our model across a wide range of evaluation test benchmarks, with PIP-MM covering the largest area in the radar chart.
In the manual assessment, we discovered that PIP-MM exhibits an exceptional capacity for precise response generation, even in the confusion mode, providing accurate replies to questions.
Furthermore, benefiting from its powerful text-visual alignment capability, PIP-MM demonstrates impressive performance while reducing half of the visual input.
Our contributions are as follows:
\begin{itemize}
    \item We propose the efficient and effective framework PIP-MM, which utilizes the off-the-shelf structures of MLLMs to realize image-text early fusion.
    \item Since our proposed framework only requires adding a text-to-image adapter, it can be applied to most MLLMs with limited training costs.
    \item Due to its strong text-visual alignment capability, PIP-MM can reduce visual input to decrease memory overhead while maintaining good generation capability.
\end{itemize}

\section{Related Work}

\subsection{\textbf{Multimodal Large Language Models}}

In recent years, research on Large Language Models (LLMs) has been extremely active \cite{zhao2023llmsurvey}, thanks to the excellent generalization of LLMs in zero-shot tasks, including GPT-3 \cite{gpt3}, PaLM \cite{chowdhery2023palm}, T5 \cite{flant5}, LLaMA \cite{llama}, GLM \cite{du2021glm}, and others.
Particularly, structures as simple and efficient as decoder-only, like GPT-3, can be easily scaled to billions of parameters, demonstrating promising patterns as model size and data increase, even exhibiting emergent capabilities as parameters scale to a certain magnitude \cite{openai2023gpt4}.
Furthermore, recent advancements in instruction tuning indicate that LLMs can be fine-tuned with limited instruction data to follow open-ended instructions in natural language \cite{instructiontuning}.
This not only significantly enhances their performance on downstream tasks but also makes them user-friendly assistants in our daily lives.

In contrast, inspired by the latest advancements in LLMs, directly training neural networks to accept multimodal inputs and produce end-to-end output responses has also proven to be feasible and promising \cite{zhu2023minigpt}.
To achieve this, an intuitive idea is to adapt LLMs to multimodal inputs by adding some additional trainable parameters and fine-tuning them on multimodal data.
For instance, Flamingo \cite{alayrac2022flamingo} is one of the early works exploring this idea.
It utilizes a visual encoder (such as CLIP-ViT \cite{clip}) to extract visual embeddings and then applies multi-layer cross-attention to integrate multimodal inputs for final prediction.
This concept has also led many researchers to follow suit, such as Qwen-VL \cite{bai2023qwen}, BLIP2 \cite{li2023blip}, InstructBLIP \cite{instructblip}, and others.
Recent efforts directly connect visual embeddings to the input of LLMs and fine-tune LLMs end-to-end \cite{bliva}.
They often add an additional projection layer to map visual embeddings to the same dimension as language embeddings and then directly input them into LLMs for further training.
Different approaches may adopt different training strategies.
For example, internLM-Xcomposer1 (internLM-XC1) \cite{internlmxcomposer} employs BERTbase equipped with cross-attention layers as a perceptual sampler between the visual encoder and LLM and additionally sets extra LORA parameters within LLM to further integrate visual and textual information internally.
Lynx \cite{zeng2024matters} utilizes a re-sampling mechanism, injecting long visual token sequences into short learnable query sequences to reduce the dimension of visual inputs.
KOSMOS-2 \cite{peng2023kosmos} does not rely on any pretrained LLMs but rather trains from scratch on a large amount of mixed data, including image-text pairs, text corpora, and interleaved image-text data.
These models are robust and demonstrate promising results in developing MLLMs.

\subsection{Prompt-aware Mechanism in MLLMs}
Although existing MLLMs leverage additional visual descriptions to extend pre-trained LLMs and demonstrate strong capabilities in image-language generation tasks, these visual descriptions are often insufficient or not perceived with respect to the prompt, resulting in ineffective descriptions.
Therefore, some works have begun to explore prompt-aware visual feature extraction to alleviate these issues.
InstructBLIP \cite{instructblip} initializes training using the pretrained BLIP-2 \cite{li2023blip} model, fine-tuning with Q-Former while keeping the image encoder and LLM frozen.
During prompt conditioning, textual prompts are not only assigned to the frozen LLM but also Q-Former, allowing it to extract prompt-aware visual features from the frozen image encoder.
BLIVA \cite{bliva} enhances the visual comprehension of textual images by utilizing visual-textual query embeddings and an additional auxiliary visual branch.
Similarly, LLaVA \cite{llava} and others also follow similar principles.
Additionally, after obtaining visual tokens through the adapter from visual features, it is currently popular to create an additional channel for fusing visual tokens with prompts by adding extra LORA parameters in the LLM \cite{dong2024internlm2}.
These are all excellent works, but due to the lack of early integration of prompt information, missing part of the image information after the image encoding stage can lead to poorer model responses.
This phenomenon becomes more pronounced under the confusion mode.

\section{Method}
\begin{figure*}[!h]
    \centering
    \includegraphics[scale=0.70]{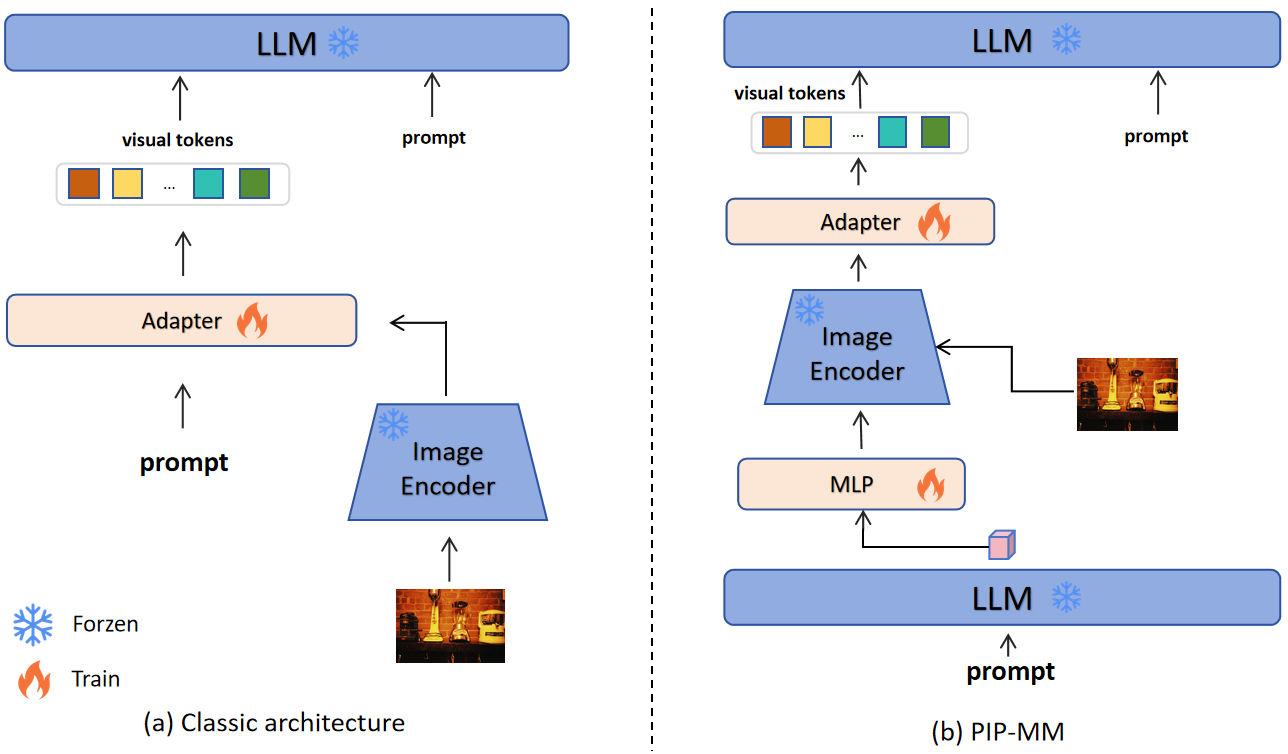}
    \caption{The comparison of the mainstream approach for integrating prompts in current MLLMs and PIP-MM.
    Classic architectures, such as InstructBLIP, do not integrate textual information during the visual encoding process; instead, they incorporate it within an Adapter, which fails to address the issue of missing visual information.
    In contrast, PIP-MM employs the inner LLM and an MLP layer to summarize the query information into a vector that replaces the image encoder's CLS token, achieving an early fusion of text and image.}
    \label{fig:architecture}
\end{figure*}

Figure \ref{fig:architecture} illustrates the differences between our model framework and current models.
Current MLLMs mainly consist of three modules: image encoder, adapter, and LLM.
In contrast, PIP-MM introduces an additional MLP.
Since PIP-MM is applicable to most existing models, considering the variability in the adapter structures of existing models whether it is Q-Former, a simple linear layer, or an MLP we refer to it collectively as the "Adapter."

\subsection{Prompt-aware Image Encoding}

Figure \ref{fig:architecture} (b) illustrates the overall architecture of PIP-MM.
Firstly, the prompt is fed into the LLM to generate a text feature vector.
To conform to the dimension settings of the image encoder and align with patch embeddings, this vector undergoes processing through an MLP to obtain T-CLS.
Subsequently, T-CLS replaces the visual CLS token and, together with patch embedding, is input into the image encoder to extract visual features.
Following this, an adapter is utilized to enhance the directive-following capability of visual features.
Finally, the LLM integrates visual information and prompt for generating the ultimate answer.

Given an image $\mathbf{I} \in \mathbb{R}^{H \times W \times C}$, it can be divided into multiple small patches $[I^1_p, I^2_p, ..., I^n_p]$,and a Prompt $X= [x_1,x_2...,x_l]$, with $l$ representing the length of the prompt, and $x_i$ representing the $i$-th token in the entire sentence.
In most cases, LLM is a classic decoder-only architecture.
At time step $t$, the state of the decoder corresponds to:
\[
\begin{aligned}
\mathbf{H}_{t} &= \text{LLM}_{\theta}(x_t|X_{<t})
\end{aligned}
\]
where LLM$_\theta$ represents the autoregressive encoding function with parameters at $\theta$.
$X_{<t}$ is the prefix of the token sequence $X$ with a length of $t-1$.

Therefore, for any given $X$, we can obtain its corresponding text hidden representation $H_l$ through LLM.
Furthermore, we convert it to $T_{class}$, namely T-CLS.
\[
\begin{aligned}
\mathbf{T}_{\text{class}} &= \text{MLP}(\mathbf{H}_{l})
\end{aligned}
\]
T-CLS is a set of feature vectors that can be directly concatenated with patch embeddings.
so we replace CLS with T-CLS to obtaining $\mathbf{z}_0$.
\[
\begin{aligned}
\mathbf{z}_0=\left[\mathbf{T}_{\text{class}};\mathbf{I}_{p}^{1} \mathbf{E} ; \mathbf{I}_{p}^{2} \mathbf{E} ; \cdots ; \mathbf{I}_{p}^{N} \mathbf{E}\right]+\mathbf{E}_{\text{pos}}
\end{aligned}
\]
Then we use $\mathbf{z}_0$ as the initial input to the ViT and achieve early interaction between text and vision within the multi-head self-attention module of the ViT.
This process results in obtaining the integrated visual-textual features, $\mathbf{z}$, at the final layer.
By inserting T-CLS, the attention weight distribution of each patch is altered, completing the initial focus of visual information and ensuring that the visual information contains as many prompt-specified objects as possible.
Most of the current prompt-aware MLLMs integrate the first fusion stage of text and images into the adapter.
As shown in Figure \ref{fig:architecture} (a), although their interaction methods may vary, a clear distinction between PIP-MM and them is whether to incorporate the prompt into the visual encoding process in advance.

\subsection{Response Generation}
After obtaining visual features $\mathbf{z}$, existing methods primarily need to generate visual tokens $\textbf{V}$ through an adapter to bridge the input and semantic differences between the image encoder and LLM.
Most models' adapter can be categorized into three types: (1) using a linear layer to directly project visual information from the latent space size of ViT to the latent space size of LLM;
(2) using a set of fixed-size trainable query vectors to query visual information based on the prompt specification through cross-attention;
(3) transforming the size of visual information through MLP and additionally inserting trainable parameters, Lora, into LLM to make visual information follow prompt requirements in Lora.
The purpose of these methods is to obtain continuous visual tokens $\textbf{V}$.
It can be input into MLLM as stably as text embeddings, so it only needs to be directly concatenated with the prompt after passing through the text embedding layer, and then the LLM can be used for autoregressive response generation.

\subsection{Two-stage Training}
After obtaining visual tokens that can be directly input into LLM, we adopt the classic two-stage training approach.
Our training objective is to find a set of model parameters that maximize the probability of generating the true answer given an input image and prompt.
\[
\begin{aligned}
\mathbf{\theta^*} = \underset{\theta}{\text{argmax}}\mathbf{\sum_i \log \text{p}_{\text{MLLM}_\theta}({A}^{(i)}|(I,P)^{(i)}; \theta)}
\end{aligned}
\]
where $\theta$ denotes the parameters of MLLM and $p_{MLLM_{\theta}}$ denotes the probability distribution entailed by these parameters.
\textbf{A}, \textbf{I}, and \textbf{P} respectively represent the true answer, image, and prompt.
The summation is over the training set and \{{$\textbf{A}^{(\text{i})}$, {$\textbf{I}^{(\text{i})}$,{$\textbf{P}^{(\text{i})}$}\} is the i-th training sample.

In the pre-training stage, we freeze almost all parameters, retaining only the gradient of the MLP part in Figure \ref{fig:architecture}(b).
In this stage, our aim is to obtain T-CLS aligned with the ViT input from the image-text pairs.
Following pre-training, although the LLM model can generate descriptions of images, it tends to produce a lot of irrelevant content when answering specific image-related questions.
To make the generation of responses smoother and of higher quality, fine-tuning in the second stage is crucial.
Compared to the training in the first stage, in terms of training parameters, we enable the adapter to train together with MLP.
Regarding training data, the text instructions in the second stage are more diverse to enhance the model's ability to follow instructions.
All training data are sourced from public datasets, and we randomly sample a portion of the data from each dataset for training, as detailed in appendix.

We use the maximum likelihood estimation algorithm to maximize the probability of the model generating data labels, i.e., cross-entropy loss based on predicting the next word given previous data labels.
For a specific sample \{{$\textbf{A}^{(\text{i})}$, {$\textbf{I}^{(\text{i})}$,{$\textbf{P}^{(\text{i})}$\}, , we obtain the visual tokens $\textbf{V}^{(\text{i})}$ corresponding to the image $\textbf{I}^{(\text{i})}$ through the image encoder and adapter.
Equation \ref{eq:xent} is equivalent to minimizing the sum of negative log-likelihoods of tokens \{$a_1,...,a_j,...a_l$\} in the answer $\textbf{A}^{(\text{i})}$ with a length $l$.
Here is the corresponding loss function formula.
\begin{equation}
\small
\begin{split}
    & \mathcal{L}_{xent} =  \\
    & - \sum_{j=1}^l \sum_a p_{\mathrm{true}} (a|Q, A_{<j}^*) \log p_{\text{MLLM}_\theta} (a |Q, A_{<j}^*; \theta) \\
\end{split}
\label{eq:xent}
\end{equation}

\begin{equation}
    \small
    Q = \text{Concat}(V,P)
\end{equation}
$Q$ represents the direct concatenation result of visual tokens and prompt text embeddings.
A denotes the prefix of sequence $A$ with a length less than $j$ \{$a_0,a_1,...a_j$\}, where $a_0$ is the predicted start symbol.
$p_{\mathrm{true}}$ represents the one-hot encoded distribution.

All data from the Image Caption section is used during the pre-training phase, while the remaining data is used for training in the second phase.
Training is conducted on 8 H800 GPUs, with the first stage requiring 52 hours and the second stage requiring 11 hours.

\section{Experiments}
In this section, we will explain the following questions through experiments, data statistics, example demonstrations, and so on: (1) How does PIP-MM perform?
(2) Can PIP-MM overcome the problems of missing visual information and sparse information density?

\subsection{Main Result}

\paragraph{\textbf{MLLM Benchmark Result}}
As shown in Table \ref{tab:main_res}, we compare our PIP-MM with state-of-the-art MLLM.
In general, we refer to PIP-MM(internLM-XC) as PIP-MM.
Here we report results in MM-Vet~\cite{yu2023mmvet}, MME~\cite{2023MMsurvey}, MMB~\cite{mmbench}, Seed-Bench~\cite{li2023seed}, MMMU~\cite{yue2023mmmu}, ScienceQA-img \cite{scienceqa}, MathVista \cite{lu2023mathvista}.
We first applied PIP-MM to the SOTA model InternLM-XC, which showed strong competitiveness in the results.
It performed slightly below the baseline model only in MathVista.
For such cases, our explanation is that most mathematical tasks focus on the global image, such as solving for unknowns when an equation appears in a picture.
Clearly, such problems do not focus on a specific part of the image, so we can only be on par with the baseline.
The reason why PIP-MM demonstrates such strong performance can be mainly attributed to two factors. Firstly, the powerful backbone provides PIP-MM with a wealth of knowledge.
Secondly, the pre-integration of text information addresses two issues existing in the original backbone, allowing the model's performance to take a step further on the original basis.
Therefore, to validate the generalization ability of PIP-MM, we chose a different backbone with weaker performance compared to InternLM-XC as the baseline.
The results were largely consistent with our expectations, and even the model trained on the weaker backbone surpassed the baseline comprehensively.
In addition to the comparative experiments with the same backbone, we also presented the results of numerous MLLMs.
Compared with many excellent models such as Qwen-VL-chat, LLAVA, etc., PIP-MM demonstrated promising prospects.

\begin{table}[!h]
\centering
\small
\begin{tabular}{ccc}
\toprule
  InternLM-XC & PIP-MM$_1$ & Comparable \\ \midrule
30.5$\%$  &43.0$\%$  &26.5$\%$  \\ 
\toprule
 Lynx & PIP-MM$_2$ & Comparable \\ \midrule
38.5$\%$  &46.0$\%$  &15.5$\%$  \\ \bottomrule
\end{tabular}
\caption{The results of the manual evaluation conducted by four volunteers on 100 confusion pattern data for both PIP-MM and the baseline model.
The subscripts 1 and 2 respectively represent the usage of InternLM-XC and Lynx as the backbone of PIP-MM.
}
\label{table:human_eval}
\end{table}

\paragraph{\textbf{Human Evaluation in Confusion Mode}}

\begin{table*}[!t]
  \begin{tabular}{lccccccc}
    \toprule
    \textbf{Model} & \textbf{MM-Vet} & \textbf{MME} &\textbf{MMB} & \textbf{Seed-Bench} & \textbf{MMMU} &\textbf{Science-img} & \textbf{MathVista} \\
    \midrule
    Flamingo \cite{alayrac2022flamingo}& 23.3& 607& 5.7 & 28.8 & 28.8 & - & 18.6 \\
    MiniGPT-4 \cite{zhu2023minigpt}& 10.5& 582& 9.4 & 29.4 & 25 & 42.84 & 22.9 \\
    VisualGLM \cite{du2021glm}& 20.3& 705& 37.6 & 47.0 & 29.9 & 45.6 & 21.5 \\
    LLAVA \cite{llava}& 32.9& \textcolor[RGB]{65,105,225}{\textbf{1631}}& \textcolor[RGB]{65,105,225}{\textbf{66.5}} & 65.8 & 35.7 & 66.8 & 25.1 \\
    InstructBLIP \cite{instructblip}& 33.1& 1212& 33.9 & 44.5 & 30.6 & 63.1 & 23.7 \\
    ShareGPT \cite{chen2023sharegpt4v}& 33.4& 1619& 67.6 & \textcolor[RGB]{65,105,225}{\textbf{69.3}} & 37.2 & 68.4 & 28.8 \\
    \rowcolor{yellow!20} Lynx \cite{zeng2024matters} & 27.6& 1373& 49.9 & 53.6 & 33.2 & - & 25.3 \\
    Monkey \cite{li2023monkey}&35.1& 1522& 59.6 & 64.3 & \textcolor[RGB]{65,105,225}{\textbf{38.9}} & \textcolor[RGB]{65,105,225}{\textbf{69.4}} & 32.5 \\
    mPLUG\_Owl2 \cite{xu2023mplug}& 37.2& 1450& 66 & 64.5 & 34.7 & 68.7 & 25.3 \\
    Qwen-VL-chat \cite{bai2023qwen}& \textcolor[RGB]{65,105,225}{\textbf{47.2}}& 1487& 61.8 & 64.8 & 37.0 & 68.2 & \textcolor[RGB]{65,105,225}{\textbf{33.8}} \\
    \rowcolor{gray!20} InternLM-XC \cite{dong2024internlm2} & \textcolor[RGB]{50,205,50}{\textbf{49.4}}& \textcolor[RGB]{50,205,50}{\textbf{1712}}& \textcolor[RGB]{50,205,50}{\textbf{80.7}} & \textcolor[RGB]{50,205,50}{\textbf{72.9}} & \textcolor[RGB]{50,205,50}{\textbf{41.4}} & \textcolor[RGB]{50,205,50}{\textbf{73.6}} & \textcolor{red}{57.9} \\
     \midrule
     \rowcolor{yellow!20} PIP-MM(Lynx) & 28.6& 1378& 50.8 & 56.4 & 34.3 & - & 25.5 \\
     \rowcolor{gray!20} PIP-MM(InternLM-XC) & \textcolor{red}{\textbf{50.8}}& \textcolor{red}{\textbf{1748}}& \textcolor{red}{\textbf{81.2}} &\textcolor{red}{\textbf{75.6}} & \textcolor{red}{\textbf{45.6}} & \textcolor{red}{\textbf{74.3}} & \textcolor[RGB]{50,205,50}{\textbf{57.2}} \\
    \bottomrule
  \end{tabular}
  \caption{The main experimental results.
  The parts with the same background color indicate that we utilized the other party's model as the backbone.
  The sections highlighted in \textcolor{red}{red}, \textcolor[RGB]{50,205,50}{green}, and \textcolor[RGB]{65,105,225}{blue} represent the first, second, and third places, respectively, under the corresponding test benchmarks.}
  \label{tab:main_res}
\end{table*}

We selected 100 images from LLaVA-150K.
For each image, we primarily posed questions regarding hidden or deceptive objects within the image, such as the concealed flag on the bus in Figure \ref{fig:intro}.
Then, we inputted each image-question pair into the model to generate responses. Subsequently, we anonymously presented the generated results from different models to four volunteers, asking them to judge which responses were good, bad, or on par.
We finally tally the number of victories of the model in each data sample and calculate the winning rate.

\subsection{Analytical Experiments}

\paragraph{\textbf{Data Ablation}}
To eliminate the possibility that the additional training data used during the training process of PIP-MM resulted in performance improvements compared to the baseline, we trained the baseline model on the same dataset to mitigate the potential impact of the data on experimental results.
As shown in Figure \ref{fig:ftbaseline}, the performance of InternLM-XC-FT, fine-tuned on the same data as PIP-MM, has decreased on all three benchmarks.
This demonstrates that the data did not influence the main conclusions of our experiments.

\begin{figure}[h]
\centering
\includegraphics[width=0.9\linewidth]{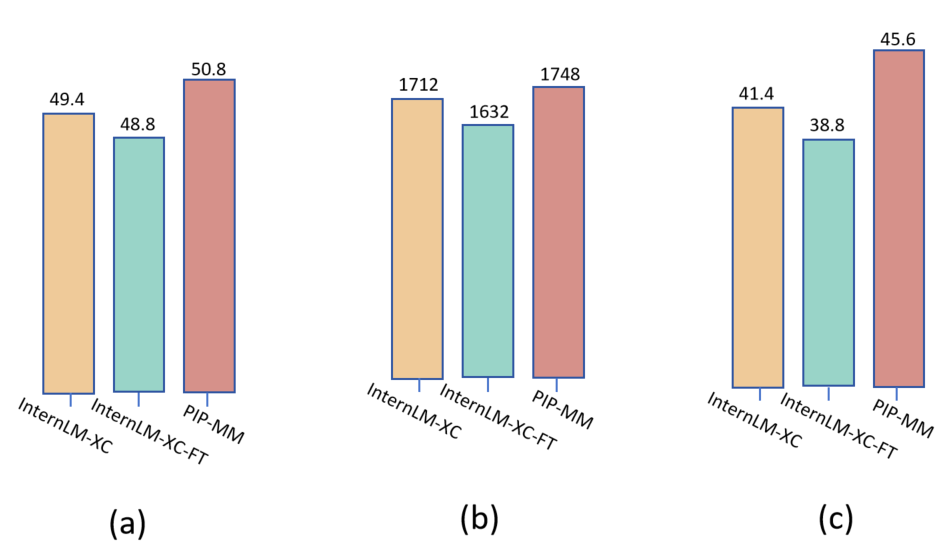}
\caption{
Elimination experiments to assess the impact of training data.
(a), (b), and (c) correspond to the results of MM-Vet, MME, and MMMU, respectively.
}
\label{fig:ftbaseline}
\end{figure}

\begin{figure*}[!h]
    \centering
    \includegraphics[scale=0.5]{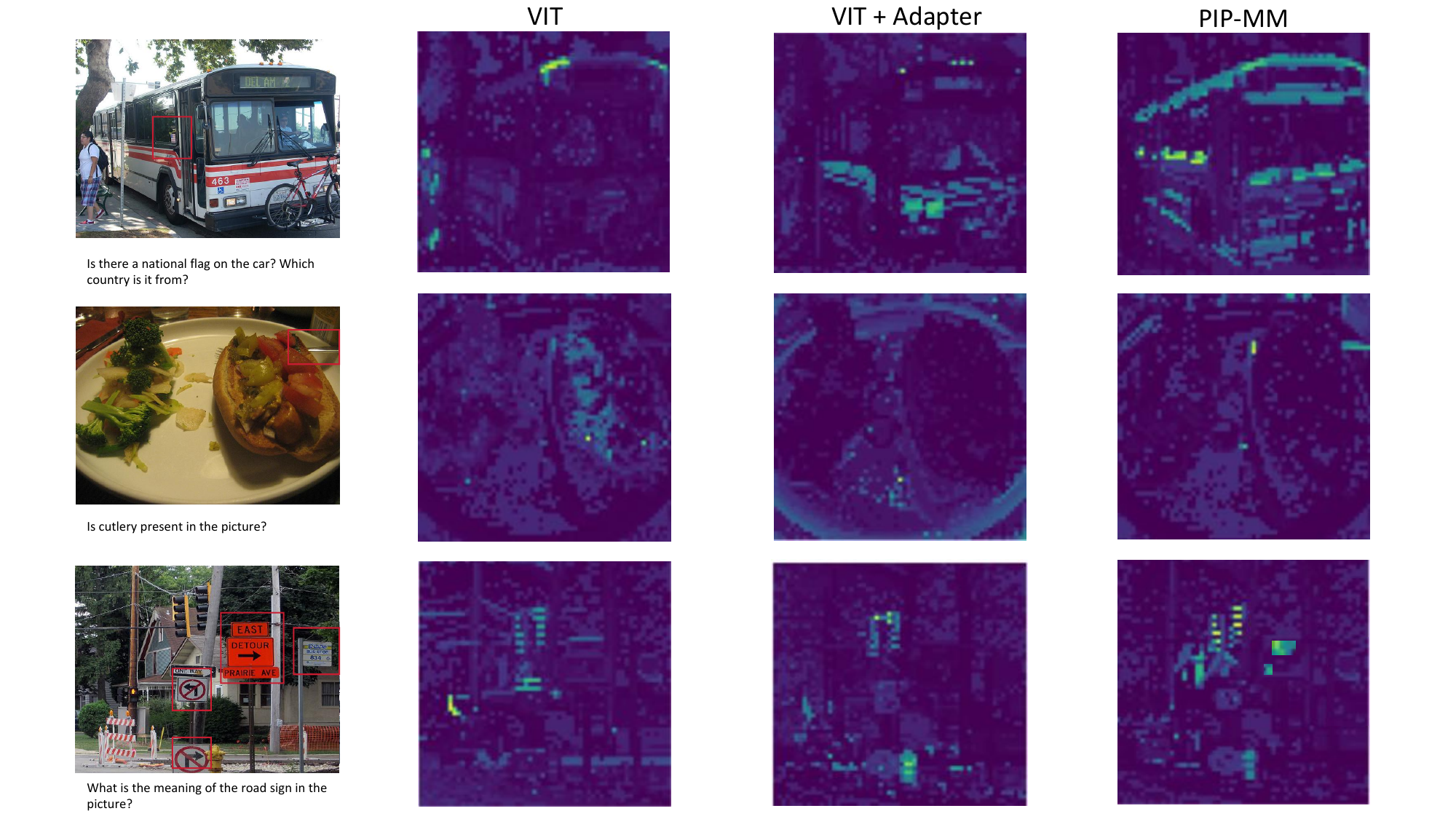}
    \caption{Attention visualization.
    The red box in the original image represents the object mentioned in the prompt.
    The highlighted part in the attention map represents the portion of visual tokens that the model focuses on describing.}
    \label{fig:att}
\end{figure*}

\begin{table}[!h]
\footnotesize
\centering
\begin{tabular}{c|ccc}
\toprule
Assessment  & Baseline & PIP-MM & PIP-MM(Ver2) \\
\midrule
MM-vet & 49.4 &  50.8 \color{red}{(+1.4)} & 49.9 \color{red}{(+0.5)}\\
\midrule
MME & 1712 & 1748 \color{red}{(+36.0)} & 1727 \color{red}{(+15.0)} \\
\midrule
MMMU & 41.4 & 45.6 \color{red}{(+4.2)} & 42.0 \color{red}{(+0.6)} \\
\bottomrule
\end{tabular}
\caption{Comparing the results of using different modules to extract textual information with InternLM-XC as the baseline.
Ver2 represents the results using the CLIP Text Encoder as the extractor.}
\label{table:instruction-ablation}
\end{table}

\paragraph{\textbf{Text Extractor Analysis}}
During the experiment, we noticed that some models inherently include a Text Encoder module, such as CLIP.
Therefore, we utilized it for text feature extraction.
Similar to PIP-MM, we employed an MLP as the text-to-image feature transformer and trained it on the same dataset.
We refer to this model as PIP-MM(Ver2).
As shown in Table \ref{table:instruction-ablation}, PIP-MM(Ver2) also demonstrated performance improvement. However, due to the powerful text summarization capability of LLM, the improvement of Ver2 unfortunately fell short of expectations.

\paragraph{\textbf{Visual Token Compression Analysis}}
As PIP-MM accomplishes an early fusion of visual and textual information, the visual token information we obtain is more focused on the objects specified by the prompt.
Therefore, we are very curious whether we can generate answers using a smaller number of visual tokens.
We tested the performance of compressed models on MM-Vet. 
As shown in Table \ref{tab:compress}, both Lynx$^{1/2}$ and InternLM-XC$^{1/2}$ were affected after compression, especially with a significant performance drop on InternLM-XC$^{1/2}$.
However, using PIP-MM showed good performance even after compression, with significant improvement, particularly when Lynx was used as the backbone.
This also confirms our experimental motivation, namely, altering the distribution of visual information through pre-integrated textual information.
It is worth mentioning that compression brings many benefits, including faster model responses and reduced memory overhead.
The specific details of the model's inference speed and memory overhead are provided in the appendix.
Additionally, in Multimodal In-Context Learning (M-ICL), shorter visual tokens enable MLLMs to support more image instances, which is particularly evident in models with very long visual tokens.

\begin{table}[!h]
\centering
\small
  \begin{tabular}{lcc}
    \toprule
    \textbf{Model} & \textbf{Visual Token Num} &  \textbf{Total} \\
    \midrule
     Lynx  &32&  27.6\\
    InternLM-XC  &1225& 49.4 \\
     \midrule
     Lynx$^{1/2}$  &16& 26.1 \\
    InternLM-XC$^{1/2}$  &512& 44.4 \\
     \midrule
     PIP-MM(Lynx)$^{1/2}$  &16& 30.0 \\
     PIP-MM(InternLM-XC)$^{1/2}$  &512& 48.7\\
    \bottomrule
  \end{tabular}
\caption{Experimental results of visual token compression on MM-Vet.
  The compression rate is 50\% .
  The superscript 1/2 represents that the model has removed 50\% of the visual tokens during the Perceive Sampler phase.
  }
 \label{tab:compress}
\end{table}

\paragraph{\textbf{Attention Visualization Analysis}}
In Figure \ref{fig:att}, we present the visual analysis of the focus of visual tokens on different parts of the image.
To increase the difficulty of the task, these data come from the confusion mode data we annotated.
The first image depicts a street scene where the object mentioned in the prompt is the flag on the side window of a bus.
From the visual features extracted by ViT, we observe that the visual information is mainly concentrated on pedestrians on the roadside, with only a small portion focusing on the bus, and even that focus is not accurate.
With the addition of Adapter, the intervention of prompt information affects the distribution of visual information, causing most of the attention to be focused on the bus, although it unfortunately does not include the window part.
In PIP-MM, since we integrate the prompt information into the visual encoding process, almost all information on the bus is encompassed by visual tokens, including the most important flag.
This explains why PIP-MM performs well in subsequent response generation.
Additionally, in the second image, the prompt specifies the fork covered by food, and in the third image, the prompt specifies the numerous confusing road signs.
We can visibly observe that PIP-MM can accurately focus on generating visual tokens based on prompt requirements.
More detailed visualization results can be found in the appendix.

\section{Conclusion}
In this paper, we propose PIP-MM, a MLLM training framework that integrates prompt information into the visual encoding process.
It can be easily applied to any existing MLLM.
To validate the effectiveness of PIP-MM, we conduct tests on multiple MLLM benchmarks and also perform manual evaluations. Both automated metrics and manual assessments demonstrate the superiority of PIP-MM.
The results of compression analysis experiments indicate that we can achieve good performance with reduced visual inputs, while speeding up model generation and reducing memory overhead.
Additionally, we enhance the credibility of PIP-MM's interpretability through statistical and visual analysis.

\bibliography{main}

\begin{thebibliography}{40}
\providecommand{\natexlab}[1]{#1}

\bibitem[{Alayrac et~al.(2022)Alayrac, Donahue, Luc, Miech, Barr, Hasson, Lenc, Mensch, Millican, Reynolds et~al.}]{alayrac2022flamingo}
Alayrac, J.-B.; Donahue, J.; Luc, P.; Miech, A.; Barr, I.; Hasson, Y.; Lenc, K.; Mensch, A.; Millican, K.; Reynolds, M.; et~al. 2022.
\newblock Flamingo: a visual language model for few-shot learning.
\newblock \emph{Advances in neural information processing systems}, 35: 23716--23736.

\bibitem[{Bai et~al.(2023)Bai, Bai, Yang, Wang, Tan, Wang, Lin, Zhou, and Zhou}]{bai2023qwen}
Bai, J.; Bai, S.; Yang, S.; Wang, S.; Tan, S.; Wang, P.; Lin, J.; Zhou, C.; and Zhou, J. 2023.
\newblock Qwen-vl: A versatile vision-language model for understanding, localization, text reading, and beyond.

\bibitem[{Brown et~al.(2020)Brown, Mann, Ryder, Subbiah, Kaplan, Dhariwal, Neelakantan, Shyam, Sastry, Askell et~al.}]{gpt3}
Brown, T.; Mann, B.; Ryder, N.; Subbiah, M.; Kaplan, J.~D.; Dhariwal, P.; Neelakantan, A.; Shyam, P.; Sastry, G.; Askell, A.; et~al. 2020.
\newblock Language models are few-shot learners.
\newblock \emph{Advances in neural information processing systems}, 33: 1877--1901.

\bibitem[{Chang et~al.(2023)Chang, Wang, Wang, Wu, Yang, Zhu, Chen, Yi, Wang, Wang et~al.}]{chang2023llmsurvey}
Chang, Y.; Wang, X.; Wang, J.; Wu, Y.; Yang, L.; Zhu, K.; Chen, H.; Yi, X.; Wang, C.; Wang, Y.; et~al. 2023.
\newblock A survey on evaluation of large language models.
\newblock \emph{ACM Transactions on Intelligent Systems and Technology}.

\bibitem[{Chen et~al.(2023)Chen, Li, Dong, Zhang, He, Wang, Zhao, and Lin}]{chen2023sharegpt4v}
Chen, L.; Li, J.; Dong, X.; Zhang, P.; He, C.; Wang, J.; Zhao, F.; and Lin, D. 2023.
\newblock Sharegpt4v: Improving large multi-modal models with better captions.
\newblock \emph{arXiv preprint arXiv:2311.12793}.

\bibitem[{Chowdhery et~al.(2023)Chowdhery, Narang, Devlin, Bosma, Mishra, Roberts, Barham, Chung, Sutton, Gehrmann et~al.}]{chowdhery2023palm}
Chowdhery, A.; Narang, S.; Devlin, J.; Bosma, M.; Mishra, G.; Roberts, A.; Barham, P.; Chung, H.~W.; Sutton, C.; Gehrmann, S.; et~al. 2023.
\newblock Palm: Scaling language modeling with pathways.
\newblock \emph{Journal of Machine Learning Research}, 24(240): 1--113.

\bibitem[{Chung et~al.(2024)Chung, Hou, Longpre, Zoph, Tay, Fedus, Li, Wang, Dehghani, Brahma et~al.}]{flant5}
Chung, H.~W.; Hou, L.; Longpre, S.; Zoph, B.; Tay, Y.; Fedus, W.; Li, Y.; Wang, X.; Dehghani, M.; Brahma, S.; et~al. 2024.
\newblock Scaling instruction-finetuned language models.
\newblock \emph{Journal of Machine Learning Research}, 25(70): 1--53.

\bibitem[{Dai et~al.(2024)Dai, Li, Li, Tiong, Zhao, Wang, Li, Fung, and Hoi}]{instructblip}
Dai, W.; Li, J.; Li, D.; Tiong, A. M.~H.; Zhao, J.; Wang, W.; Li, B.; Fung, P.~N.; and Hoi, S. 2024.
\newblock Instructblip: Towards general-purpose vision-language models with instruction tuning.
\newblock \emph{Advances in Neural Information Processing Systems}, 36.

\bibitem[{Dong et~al.(2024)Dong, Zhang, Zang, Cao, Wang, Ouyang, Wei, Zhang, Duan, Cao et~al.}]{dong2024internlm2}
Dong, X.; Zhang, P.; Zang, Y.; Cao, Y.; Wang, B.; Ouyang, L.; Wei, X.; Zhang, S.; Duan, H.; Cao, M.; et~al. 2024.
\newblock InternLM-XComposer2: Mastering free-form text-image composition and comprehension in vision-language large model.
\newblock \emph{arXiv preprint arXiv:2401.16420}.

\bibitem[{Du et~al.(2021)Du, Qian, Liu, Ding, Qiu, Yang, and Tang}]{du2021glm}
Du, Z.; Qian, Y.; Liu, X.; Ding, M.; Qiu, J.; Yang, Z.; and Tang, J. 2021.
\newblock Glm: General language model pretraining with autoregressive blank infilling.
\newblock \emph{arXiv preprint arXiv:2103.10360}.

\bibitem[{Han et~al.(2022)Han, Wang, Chen, Chen, Guo, Liu, Tang, Xiao, Xu, Xu et~al.}]{vit2022survey}
Han, K.; Wang, Y.; Chen, H.; Chen, X.; Guo, J.; Liu, Z.; Tang, Y.; Xiao, A.; Xu, C.; Xu, Y.; et~al. 2022.
\newblock A survey on vision transformer.
\newblock \emph{IEEE transactions on pattern analysis and machine intelligence}, 45(1): 87--110.

\bibitem[{Hu et~al.(2024)Hu, Xu, Li, Li, Chen, and Tu}]{bliva}
Hu, W.; Xu, Y.; Li, Y.; Li, W.; Chen, Z.; and Tu, Z. 2024.
\newblock Bliva: A simple multimodal llm for better handling of text-rich visual questions.
\newblock In \emph{Proceedings of the AAAI Conference on Artificial Intelligence}, volume~38, 2256--2264.

\bibitem[{Kirillov et~al.(2023)Kirillov, Mintun, Ravi, Mao, Rolland, Gustafson, Xiao, Whitehead, Berg, Lo, Doll{\'a}r, and Girshick}]{vitsam}
Kirillov, A.; Mintun, E.; Ravi, N.; Mao, H.; Rolland, C.; Gustafson, L.; Xiao, T.; Whitehead, S.; Berg, A.~C.; Lo, W.-Y.; Doll{\'a}r, P.; and Girshick, R. 2023.
\newblock Segment Anything.
\newblock \emph{arXiv:2304.02643}.

\bibitem[{Li et~al.(2023{\natexlab{a}})Li, Wang, Wang, Ge, Ge, and Shan}]{li2023seed}
Li, B.; Wang, R.; Wang, G.; Ge, Y.; Ge, Y.; and Shan, Y. 2023{\natexlab{a}}.
\newblock Seed-bench: Benchmarking multimodal llms with generative comprehension.
\newblock \emph{arXiv preprint arXiv:2307.16125}.

\bibitem[{Li et~al.(2023{\natexlab{b}})Li, Li, Savarese, and Hoi}]{li2023blip}
Li, J.; Li, D.; Savarese, S.; and Hoi, S. 2023{\natexlab{b}}.
\newblock Blip-2: Bootstrapping language-image pre-training with frozen image encoders and large language models.
\newblock In \emph{International conference on machine learning}, 19730--19742. PMLR.

\bibitem[{Li et~al.(2023{\natexlab{c}})Li, Yang, Liu, Ma, Zhang, Yang, Sun, Liu, and Bai}]{li2023monkey}
Li, Z.; Yang, B.; Liu, Q.; Ma, Z.; Zhang, S.; Yang, J.; Sun, Y.; Liu, Y.; and Bai, X. 2023{\natexlab{c}}.
\newblock Monkey: Image resolution and text label are important things for large multi-modal models.
\newblock \emph{arXiv preprint arXiv:2311.06607}.

\bibitem[{Liu et~al.(2023{\natexlab{a}})Liu, Li, Li, and Lee}]{llava}
Liu, H.; Li, C.; Li, Y.; and Lee, Y.~J. 2023{\natexlab{a}}.
\newblock Improved baselines with visual instruction tuning.
\newblock \emph{arXiv preprint arXiv:2310.03744}.

\bibitem[{Liu et~al.(2023{\natexlab{b}})Liu, Duan, Zhang, Li, Zhang, Zhao, Yuan, Wang, He, Liu et~al.}]{mmbench}
Liu, Y.; Duan, H.; Zhang, Y.; Li, B.; Zhang, S.; Zhao, W.; Yuan, Y.; Wang, J.; He, C.; Liu, Z.; et~al. 2023{\natexlab{b}}.
\newblock Mmbench: Is your multi-modal model an all-around player?
\newblock \emph{arXiv preprint arXiv:2307.06281}.

\bibitem[{Lu et~al.(2023)Lu, Bansal, Xia, Liu, Li, Hajishirzi, Cheng, Chang, Galley, and Gao}]{lu2023mathvista}
Lu, P.; Bansal, H.; Xia, T.; Liu, J.; Li, C.; Hajishirzi, H.; Cheng, H.; Chang, K.-W.; Galley, M.; and Gao, J. 2023.
\newblock Mathvista: Evaluating mathematical reasoning of foundation models in visual contexts.
\newblock \emph{arXiv preprint arXiv:2310.02255}.

\bibitem[{Lu et~al.(2022)Lu, Mishra, Xia, Qiu, Chang, Zhu, Tafjord, Clark, and Kalyan}]{scienceqa}
Lu, P.; Mishra, S.; Xia, T.; Qiu, L.; Chang, K.-W.; Zhu, S.-C.; Tafjord, O.; Clark, P.; and Kalyan, A. 2022.
\newblock Learn to Explain: Multimodal Reasoning via Thought Chains for Science Question Answering.
\newblock In \emph{The 36th Conference on Neural Information Processing Systems (NeurIPS)}.

\bibitem[{Marino et~al.(2019)Marino, Rastegari, Farhadi, and Mottaghi}]{okvqa}
Marino, K.; Rastegari, M.; Farhadi, A.; and Mottaghi, R. 2019.
\newblock Ok-vqa: A visual question answering benchmark requiring external knowledge.
\newblock In \emph{Proceedings of the IEEE/cvf conference on computer vision and pattern recognition}, 3195--3204.

\bibitem[{Mathew, Karatzas, and Jawahar(2021)}]{docvqa}
Mathew, M.; Karatzas, D.; and Jawahar, C. 2021.
\newblock Docvqa: A dataset for vqa on document images.
\newblock In \emph{Proceedings of the IEEE/CVF winter conference on applications of computer vision}, 2200--2209.

\bibitem[{OpenAI(2023{\natexlab{a}})}]{gpt4}
OpenAI, R. 2023{\natexlab{a}}.
\newblock GPT-4 technical report.
\newblock \emph{arXiv}, 2303--08774.

\bibitem[{OpenAI(2023{\natexlab{b}})}]{openai2023gpt4}
OpenAI, R. 2023{\natexlab{b}}.
\newblock GPT-4 technical report.
\newblock \emph{arXiv}, 2303--08774.

\bibitem[{Peng et~al.(2023{\natexlab{a}})Peng, Li, He, Galley, and Gao}]{instructiontuning}
Peng, B.; Li, C.; He, P.; Galley, M.; and Gao, J. 2023{\natexlab{a}}.
\newblock Instruction tuning with gpt-4.
\newblock \emph{arXiv preprint arXiv:2304.03277}.

\bibitem[{Peng et~al.(2023{\natexlab{b}})Peng, Wang, Dong, Hao, Huang, Ma, and Wei}]{peng2023kosmos}
Peng, Z.; Wang, W.; Dong, L.; Hao, Y.; Huang, S.; Ma, S.; and Wei, F. 2023{\natexlab{b}}.
\newblock Kosmos-2: Grounding multimodal large language models to the world.
\newblock \emph{arXiv preprint arXiv:2306.14824}.

\bibitem[{Radford et~al.(2021)Radford, Kim, Hallacy, Ramesh, Goh, Agarwal, Sastry, Askell, Mishkin, Clark et~al.}]{clip}
Radford, A.; Kim, J.~W.; Hallacy, C.; Ramesh, A.; Goh, G.; Agarwal, S.; Sastry, G.; Askell, A.; Mishkin, P.; Clark, J.; et~al. 2021.
\newblock Learning transferable visual models from natural language supervision.
\newblock In \emph{International conference on machine learning}, 8748--8763. PMLR.

\bibitem[{Sharma et~al.(2018)Sharma, Ding, Goodman, and Soricut}]{cc3m}
Sharma, P.; Ding, N.; Goodman, S.; and Soricut, R. 2018.
\newblock Conceptual captions: A cleaned, hypernymed, image alt-text dataset for automatic image captioning.
\newblock In \emph{Proceedings of the 56th Annual Meeting of the Association for Computational Linguistics (Volume 1: Long Papers)}, 2556--2565.

\bibitem[{Touvron et~al.(2023)Touvron, Martin, Stone, Albert, Almahairi, Babaei, Bashlykov, Batra, Bhargava, Bhosale et~al.}]{llama}
Touvron, H.; Martin, L.; Stone, K.; Albert, P.; Almahairi, A.; Babaei, Y.; Bashlykov, N.; Batra, S.; Bhargava, P.; Bhosale, S.; et~al. 2023.
\newblock Llama 2: Open foundation and fine-tuned chat models.
\newblock \emph{arXiv preprint arXiv:2307.09288}.

\bibitem[{Wang et~al.(2023)Wang, Lv, Yu, Hong, Qi, Wang, Ji, Yang, Zhao, Song et~al.}]{wang2023cogvlm}
Wang, W.; Lv, Q.; Yu, W.; Hong, W.; Qi, J.; Wang, Y.; Ji, J.; Yang, Z.; Zhao, L.; Song, X.; et~al. 2023.
\newblock Cogvlm: Visual expert for pretrained language models.
\newblock \emph{arXiv preprint arXiv:2311.03079}.

\bibitem[{Wu et~al.(2020)Wu, Xu, Dai, Wan, Zhang, Yan, Tomizuka, Gonzalez, Keutzer, and Vajda}]{googlevit}
Wu, B.; Xu, C.; Dai, X.; Wan, A.; Zhang, P.; Yan, Z.; Tomizuka, M.; Gonzalez, J.; Keutzer, K.; and Vajda, P. 2020.
\newblock Visual Transformers: Token-based Image Representation and Processing for Computer Vision.
\newblock arXiv:2006.03677.

\bibitem[{Xu et~al.(2023)Xu, Ye, Yan, Shi, Ye, Xu, Li, Bi, Qian, Wang et~al.}]{xu2023mplug}
Xu, H.; Ye, Q.; Yan, M.; Shi, Y.; Ye, J.; Xu, Y.; Li, C.; Bi, B.; Qian, Q.; Wang, W.; et~al. 2023.
\newblock mplug-2: A modularized multi-modal foundation model across text, image and video.
\newblock In \emph{International Conference on Machine Learning}, 38728--38748. PMLR.

\bibitem[{Yin et~al.(2023)Yin, Fu, Zhao, Li, Sun, Xu, and Chen}]{2023MMsurvey}
Yin, S.; Fu, C.; Zhao, S.; Li, K.; Sun, X.; Xu, T.; and Chen, E. 2023.
\newblock A survey on multimodal large language models.
\newblock \emph{arXiv preprint arXiv:2306.13549}.

\bibitem[{Yu et~al.(2023)Yu, Yang, Li, Wang, Lin, Liu, Wang, and Wang}]{yu2023mmvet}
Yu, W.; Yang, Z.; Li, L.; Wang, J.; Lin, K.; Liu, Z.; Wang, X.; and Wang, L. 2023.
\newblock Mm-vet: Evaluating large multimodal models for integrated capabilities.
\newblock \emph{arXiv preprint arXiv:2308.02490}.

\bibitem[{Yue et~al.(2023)Yue, Ni, Zhang, Zheng, Liu, Zhang, Stevens, Jiang, Ren, Sun et~al.}]{yue2023mmmu}
Yue, X.; Ni, Y.; Zhang, K.; Zheng, T.; Liu, R.; Zhang, G.; Stevens, S.; Jiang, D.; Ren, W.; Sun, Y.; et~al. 2023.
\newblock Mmmu: A massive multi-discipline multimodal understanding and reasoning benchmark for expert agi.
\newblock \emph{arXiv preprint arXiv:2311.16502}.

\bibitem[{Zeng et~al.(2024)Zeng, Zhang, Zheng, Xia, Wei, Wei, Zhang, Kong, and Song}]{zeng2024matters}
Zeng, Y.; Zhang, H.; Zheng, J.; Xia, J.; Wei, G.; Wei, Y.; Zhang, Y.; Kong, T.; and Song, R. 2024.
\newblock What Matters in Training a GPT4-Style Language Model with Multimodal Inputs?
\newblock In \emph{Proceedings of the 2024 Conference of the North American Chapter of the Association for Computational Linguistics: Human Language Technologies (Volume 1: Long Papers)}, 7930--7957.

\bibitem[{Zhang et~al.(2023)Zhang, Dong, Wang, Cao, Xu, Ouyang, Zhao, Ding, Zhang, Duan, Zhang, Yan, Zhang, Li, Li, Chen, He, Zhang, Qiao, Lin, and Wang}]{internlmxcomposer}
Zhang, P.; Dong, X.; Wang, B.; Cao, Y.; Xu, C.; Ouyang, L.; Zhao, Z.; Ding, S.; Zhang, S.; Duan, H.; Zhang, W.; Yan, H.; Zhang, X.; Li, W.; Li, J.; Chen, K.; He, C.; Zhang, X.; Qiao, Y.; Lin, D.; and Wang, J. 2023.
\newblock InternLM-XComposer: A Vision-Language Large Model for Advanced Text-image Comprehension and Composition.
\newblock \emph{arXiv preprint arXiv:2309.15112}.

\bibitem[{Zhao et~al.(2023)Zhao, Zhou, Li, Tang, Wang, Hou, Min, Zhang, Zhang, Dong et~al.}]{zhao2023llmsurvey}
Zhao, W.~X.; Zhou, K.; Li, J.; Tang, T.; Wang, X.; Hou, Y.; Min, Y.; Zhang, B.; Zhang, J.; Dong, Z.; et~al. 2023.
\newblock A survey of large language models.
\newblock \emph{arXiv preprint arXiv:2303.18223}.

\bibitem[{Zheng et~al.(2024)Zheng, Chiang, Sheng, Zhuang, Wu, Zhuang, Lin, Li, Li, Xing et~al.}]{vicuna}
Zheng, L.; Chiang, W.-L.; Sheng, Y.; Zhuang, S.; Wu, Z.; Zhuang, Y.; Lin, Z.; Li, Z.; Li, D.; Xing, E.; et~al. 2024.
\newblock Judging llm-as-a-judge with mt-bench and chatbot arena.
\newblock \emph{Advances in Neural Information Processing Systems}, 36.

\bibitem[{Zhu et~al.(2023)Zhu, Chen, Shen, Li, and Elhoseiny}]{zhu2023minigpt}
Zhu, D.; Chen, J.; Shen, X.; Li, X.; and Elhoseiny, M. 2023.
\newblock Minigpt-4: Enhancing vision-language understanding with advanced large language models.
\newblock \emph{arXiv preprint arXiv:2304.10592}.

\end{thebibliography}


\end{document}


\maketitle

\section{Background of VIT}

Because our method primarily focuses on the visual encoding part, and most MLLM's visual encoders adopt the ViT architecture, we need to introduce ViT in this section.
Additionally, By analyzing the differences between CLS in pre-training tasks and visual-language tasks, we explain the feasibility of CLS replacement schemes.
\begin{figure}[h]
\centering
\includegraphics[width=0.9\linewidth]{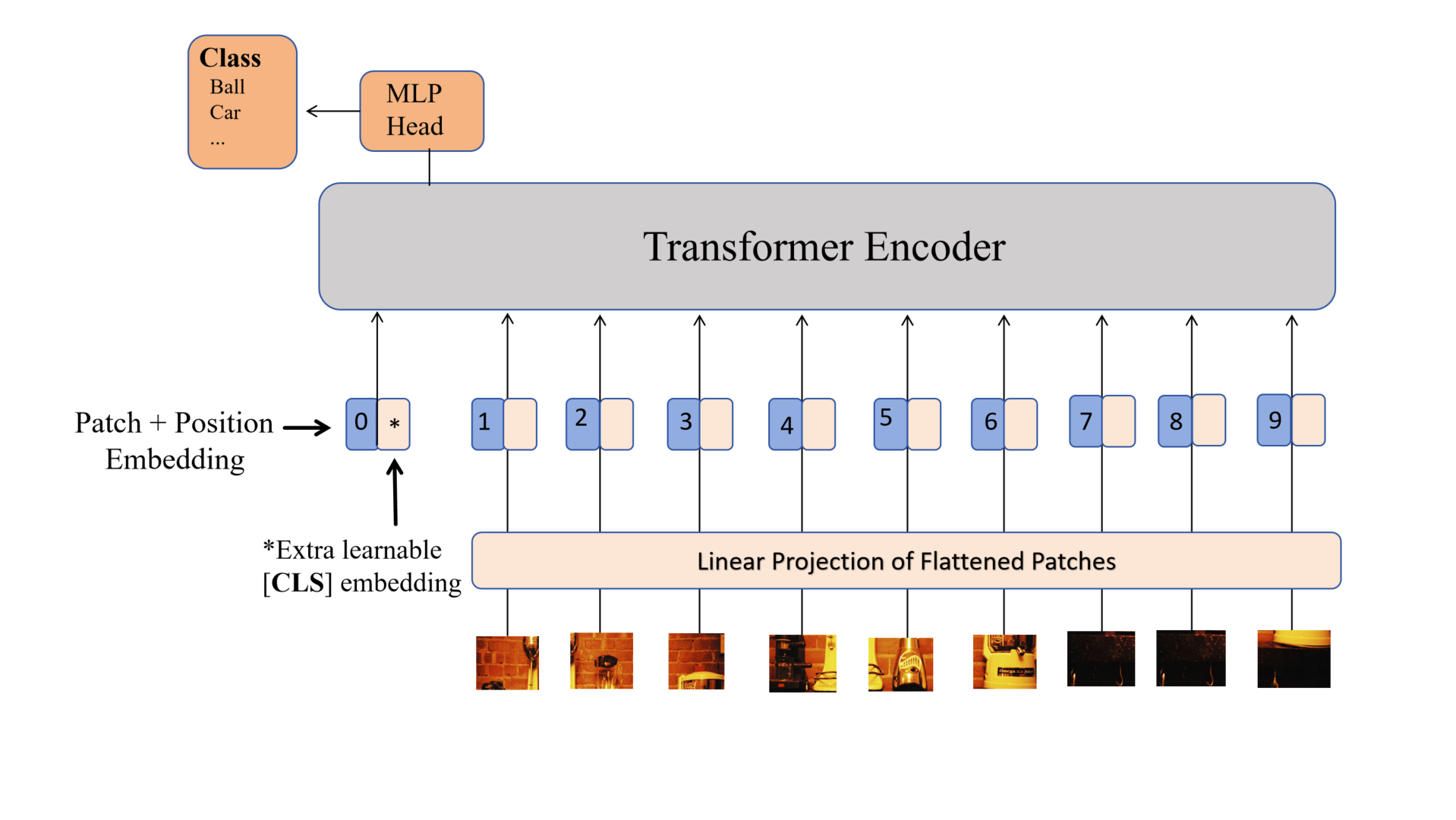}
\caption{
The schematic diagram of ViT, where CLS is regarded as the global feature of the image and is used for image classification prediction.
}
\label{fig:VIT}
\end{figure}

Figure \ref{fig:VIT} illustrates the existing workflow of ViT.
ViT transforms a 2D image into a 1D token embeddings sequence to feed into a standard Transformer network.
ViT divides the entire image $\mathbf{I} \in \mathbb{R}^{H \times W \times C}$ into a sequence of 2D patches $\mathbf{I}_{p} \in \mathbb{R}^{N \times\left(P^{2} \cdot C\right)}$, where $(H,W)$ represents the pixel resolution of the original image, $C$ represents the number of channels in the image (usually 3), $P \times P$ indicates the resolution of each image patch (typically 16), and $N=H \times W / P^{2}$ is the total number of patches (i.e., the input sequence length).
Subsequently, ViT employs a learnable linear transformation to map each patch to a latent $D$-dimensional embedding.
To encode spatial information for each patch, ViT adds learnable 1D position embeddings to the patch embeddings to preserve positional information.
Additionally, for the convenience of image classification tasks, a learnable embedding of the CLS token ($\mathbf{z}_{0}^{0} = \mathbf{I}_{\text{class}}$) is added to the beginning of the sequence of embedded image patches.
The final representation of the entire sequence is as follows:
\begin{align}
\mathbf{z}_{0}=\left[\mathbf{I}_{\text{class}};\mathbf{I}_{p}^{1} \mathbf{E} ; \mathbf{I}_{p}^{2} \mathbf{E} ; \cdots ; \mathbf{I}_{p}^{N} \mathbf{E}\right]+\mathbf{E}_{\text{pos}}
\label{align:eq1}
\end{align}
where $\mathbf{E} \in \mathbb{R}^{\left(P^{2} \cdot C\right) \times D}$ is the patch embedding projection, and $\mathbf{E}_{\text{pos}} \in \mathbb{R}^{(N+1) \times D}$ represents the position embeddings.

The Transformer encoder consists of $L$ layers of multi-head self-attention (MSA) and multi-layer perceptron (MLP) blocks.
Layer normalization (LN) is applied before every block, and residual connections are applied after every block.
The MLP contains two layers with a GELU non-linearity.
Therefore, the output of the $i$-th layer can be written as follows:

\begin{align}
\mathbf{z}_{\ell}^{\prime} &= \text{MSA}\left(\text{LN}\left(\mathbf{z}_{\ell-1}\right)\right)+\mathbf{z}_{\ell-1} \label{align:eq2} \\  
\mathbf{z}_{\ell} &= \text{MLP}\left(\text{LN}\left(\mathbf{z}_{\ell}^{\prime}\right)\right)+\mathbf{z}_{\ell}^{\prime} \label{align:eq3}
\end{align}

The final encoded image is $\mathbf{z}_{L}$, where the CLS token $\mathbf{z}_{L}^0$ final state serves as the overall representation of the entire image, used for image classification.
In ViT pre-training tasks, a classification head implemented by a multi-layer perceptron with hidden layers attaches to $\mathbf{z}_{L}^0$, followed by using cross-entropy loss to supervise the prediction results of classification.
This forces the CLS token to learn the overall semantics of the image.

\begin{figure}[h]
\centering
\includegraphics[width=0.9\linewidth]{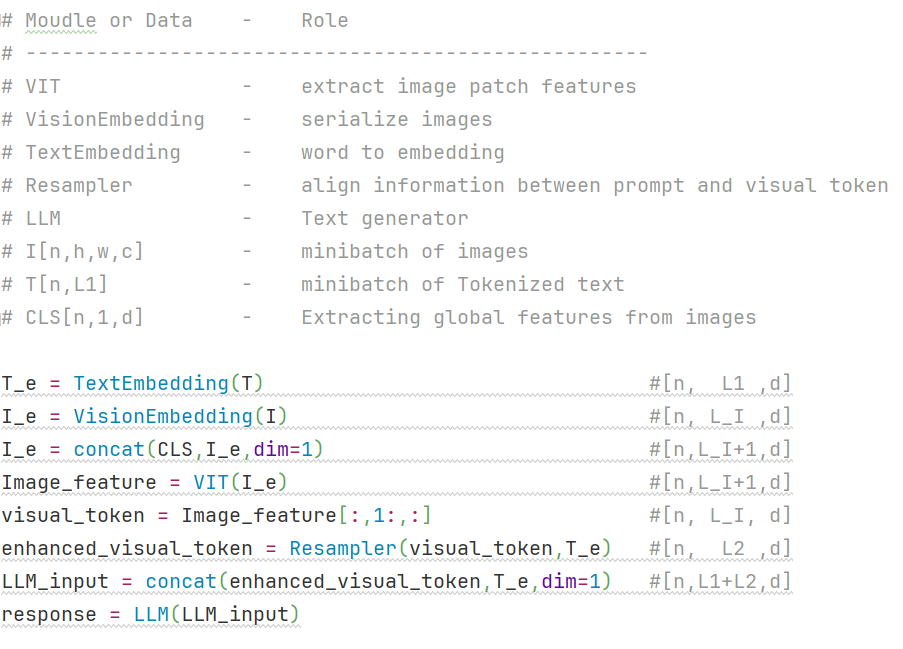}
\caption{
In the visual-language task, the pseudocode for generating model responses.
The diagram contains the data flow trajectory of CLS within the model.
}
\label{fig:code}
\end{figure}

It is undeniable that CLS plays a crucial role in traditional computer vision tasks and pre-training tasks such as image-text matching.
However, in vision-language tasks, the role of CLS is minimal.
In vision-language tasks, we typically refer to the non-CLS part of ViT encoding (i.e., all patch features) as visual tokens.
After obtaining visual tokens, most existing methods align visual tokens with the prompt input in a visual resampler to align information between vision and text, thereby enhancing the information density of prompt-specified objects in visual tokens.
Finally, the enhanced visual tokens are concatenated with text embeddings and directly input into the LLM to generate responses.
As shown in Figure \ref{fig:code}, before generating visual tokens, the information at the first position of the image features, namely the CLS token, was excluded.
It is evident that CLS does not participate in meaningful computations in vision-language tasks, providing an opportunity for PIP-MM research.

\section{Hyperparameter Experiment}

Our model simply added a text-to-image adapter.
Below, we will showcase the experimental results of this adapter under different configurations. The superscript numbers following the adapter names represent the number of layers stacked in the module.
Additionally, we also tabulated the corresponding module parameters.
As shown in Table \ref{tab:hpE}, using a single-layer linear layer resulted in a significant performance drop, likely because the process of transforming textual features into visual embeddings is complex, and a simple linear projection cannot capture this complexity.
Therefore, we replaced it with an MLP module with a larger parameter count and nonlinear transformations. We found that the performance peaked when stacking 4 layers.

\begin{table}[!h]
\centering
\caption{Using different modules as the text-to-image adapter, we evaluate the model performance on MM-Vet and MME to validate the quality of hyperparameters.
The superscript of the adapter indicates the number of layers stacked in the module.}
\setlength{\tabcolsep}{8pt}
\begin{tabular}{lcccc}
\toprule
\textbf{Adapter} & \textbf{Parameters} & \textbf{MM-Vet} &\textbf{ MME} \\
\midrule
Linear & 4M & 46.0 & 1578 \\
MLP$^{2}$ & 21M & 48.8 & 1632 \\
MLP$^{3}$ & 38M & 49.9 & 1689 \\
MLP$^{4}$ & 55M & 50.8 & 1748 \\
MLP$^{5}$ & 71M & 50.5 & 1703 \\
\bottomrule
\end{tabular}
\label{tab:hpE}
\end{table}

\section{Memory Cost and Inference Speed}
In our compression analysis experiment, we demonstrated the compression capability of PIP-MM, showing that our model performs very well even after removing half of the tokens.
Although it did not show significant improvement, we further demonstrate the advantages of compression in other aspects, such as memory consumption and inference speed.
As shown in Table \ref{tab:M&S}, compressing half of the visual tokens saved close to 5GB of memory. Additionally, we tested the inference speed on the Nvidia H800 device using MM-Vet. On a test set of size 218, we reduced the time by over two minutes.
Undeniably, our model, without compression, incurs slightly higher memory and time overhead due to the addition of an extra MLP, and the need for each visual encoding to undergo an LLM encoding before.
However, the impact is not very significant, and the substantial improvement in the generated results justifies these minor sacrifices.

\begin{table}[!h]
\caption{Testing the memory consumption and speed after compressing visual tokens on MM-Vet.}
\centering
\begin{tabular}{c|cc}
\toprule
Model  & Memory & Time/s  \\
\midrule
InternLM-XC & 40529MB & 963 \\
\midrule
PIP-MM & 40835MB & 990   \\
\midrule
PIP-MM$^{1/2}$ & 35685MB & 776  \\
\bottomrule
\end{tabular}
\label{tab:M&S}
\end{table}

\section{Training Data}
In this section, we introduced the data sources used to train our model.
Among them, shareGPT, COCO Caption, and cc3m are caption datasets, while the rest of the data are high-quality VQA datasets.
\begin{table}[h]
    \centering
    \scalebox{0.9}{
    \begin{tabular}{c|c}
    \toprule 
    Dataset          & Samples \\ \midrule 
    ShareGPT \cite{chen2023sharegpt4v} & 40K    \\
    COCO Caption \cite{cococaption}    & 80K     \\
    cc3m  \cite{cc3m}           & 1.5M    \\ \midrule 
    VQAV2      \cite{vqav2}      & 100K    \\
    TextVQA      \cite{textvqa}  & 34K    \\
    OKVQA       \cite{okvqa}    & 15K     \\ 
    LLaVA-150k  \cite{llava150k}       & 100K     \\ 
    MathQA   \cite{2019-mathqa}    & 20K    \\ 
    DocVQA     \cite{docvqa}    & 50K  \\
    TabFact    \cite{tabfact}     & 50K    \\
    InfoVQA  \cite{infovqa}        & 20K     \\ \midrule 
    Total                                     & 2.00M   \\ 
    \bottomrule
    \end{tabular}}
    \label{tab:data}
    \caption{The detailed information of the training data for PIP-MM is entirely sourced from publicly available datasets.
    }
\end{table}

\section{Attention Visualization}
In this section, we will present more attention visualizations.
We can observe the effectiveness of different strategies by focusing on the attention maps' positions specified by the Prompt for different objects and various strategies.
From the examples below, it can be seen that using a prompt-aware image encoding method makes the visual information input to the LLM more focused on the specified part of the prompt, thus achieving precise generation results.

\begin{figure*}[!t]
    \centering
    \includegraphics[scale=0.49]{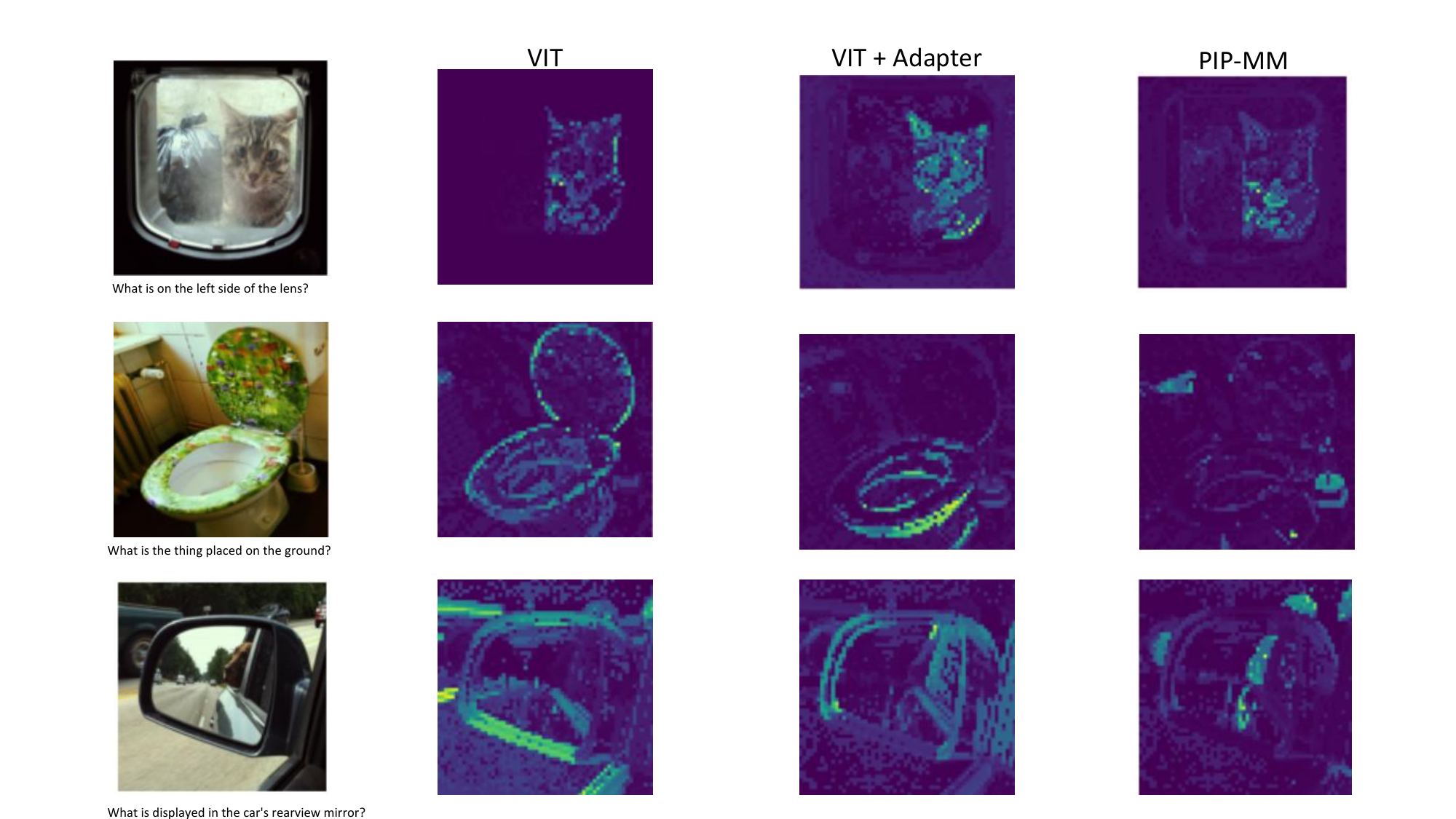}
    \label{fig:ATT1}
\vspace{3mm}
\end{figure*}

\begin{figure*}[!t]
    \centering
    \includegraphics[scale=0.50]{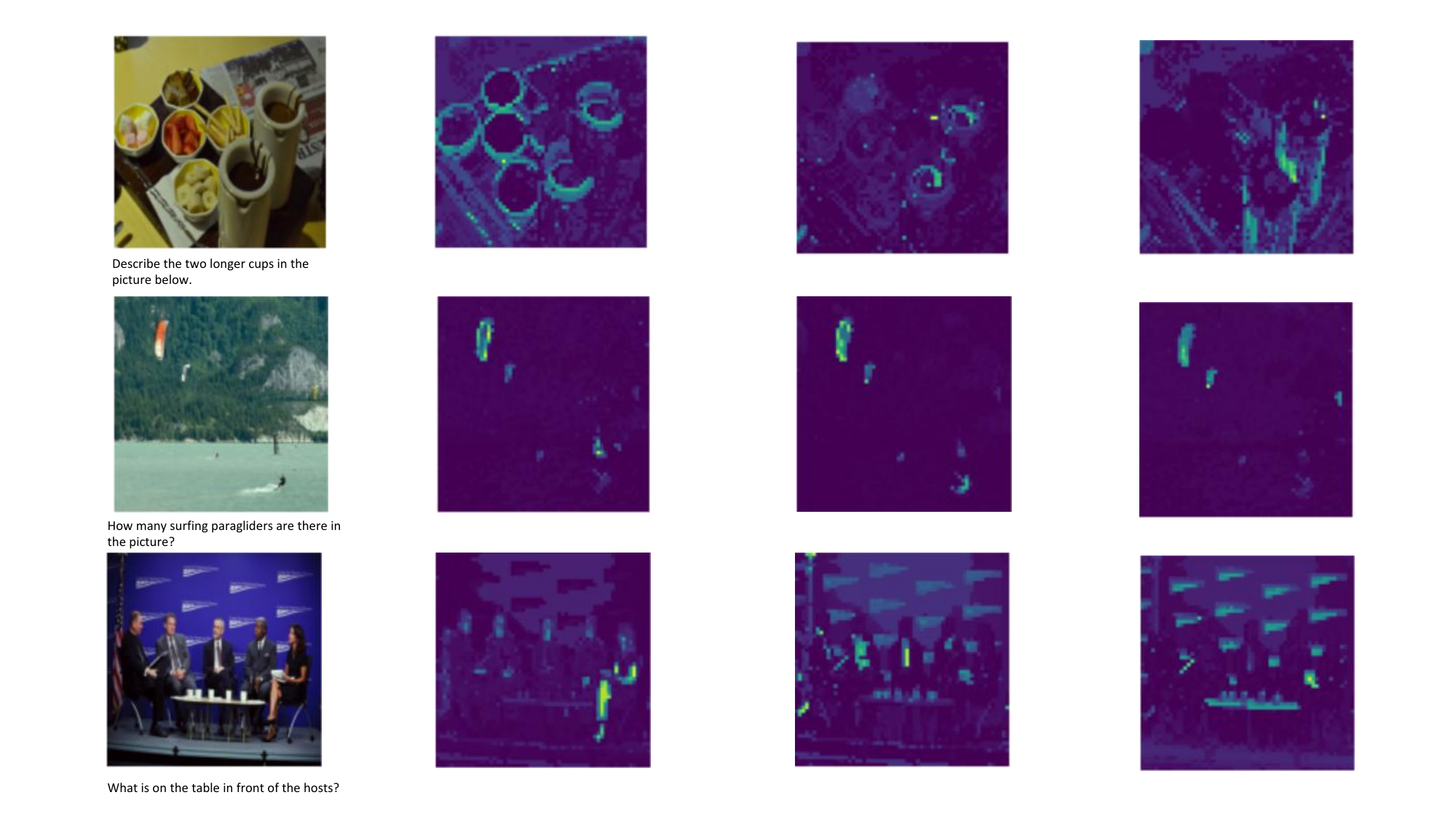}
    \label{fig:ATT2}
\end{figure*}

\newpage
\bibliography{main}
\newpage
\clearpage
\paragraph{This paper:}
\begin{itemize}
    \item Includes a conceptual outline and/or pseudocode description of AI methods introduced (yes/partial/no/NA){\bf yes}
    \item Clearly delineates statements that are opinions, hypothesis, and speculation from objective facts and results (yes/no){\bf yes}
    \item Provides well marked pedagogical references for less-familiare readers to gain background necessary to replicate the paper (yes/no){\bf yes}
\end{itemize}

\paragraph{Does this paper make theoretical contributions? (yes/no)}
{\bf yes}

If yes, please complete the list below.

\begin{itemize}
    \item All assumptions and restrictions are stated clearly and formally. (yes/partial/no){\bf yes}
    \item All novel claims are stated formally (e.g., in theorem statements). (yes/partial/no){\bf yes}
    \item Proofs of all novel claims are included. (yes/partial/no){\bf yes}
    \item Proof sketches or intuitions are given for complex and/or novel results. (yes/partial/no){\bf yes}
    \item Appropriate citations to theoretical tools used are given. (yes/partial/no){\bf yes}
    \item All theoretical claims are demonstrated empirically to hold. (yes/partial/no/NA){\bf NA}
    \item All experimental code used to eliminate or disprove claims is included. (yes/no/NA){\bf NA}
\end{itemize}

\paragraph{Does this paper rely on one or more datasets? (yes/no)}
{\bf no}

If yes, please complete the list below.

\begin{itemize}
    \item A motivation is given for why the experiments are conducted on the selected datasets (yes/partial/no/NA)
    \item All novel datasets introduced in this paper are included in a data appendix. (yes/partial/no/NA)
    \item All novel datasets introduced in this paper will be made publicly available upon publication of the paper with a license that allows free usage for research purposes. (yes/partial/no/NA)
    \item All datasets drawn from the existing literature (potentially including authors’ own previously published work) are accompanied by appropriate citations. (yes/no/NA)
    \item All datasets drawn from the existing literature (potentially including authors’ own previously published work) are publicly available. (yes/partial/no/NA)
    \item All datasets that are not publicly available are described in detail, with explanation why publicly available alternatives are not scientifically satisficing. (yes/partial/no/NA)
\end{itemize}

\paragraph{Does this paper include computational experiments? (yes/no)}
{\bf no}

If yes, please complete the list below.

\begin{itemize}
\item Any code required for pre-processing data is included in the appendix. (yes/partial/no).
\item All source code required for conducting and analyzing the experiments is included in a code appendix. (yes/partial/no)
\item All source code required for conducting and analyzing the experiments will be made publicly available upon publication of the paper with a license that allows free usage for research purposes. (yes/partial/no)
\item All source code implementing new methods have comments detailing the implementation, with references to the paper where each step comes from (yes/partial/no)
\item If an algorithm depends on randomness, then the method used for setting seeds is described in a way sufficient to allow replication of results. (yes/partial/no/NA)
\item This paper specifies the computing infrastructure used for running experiments (hardware and software), including GPU/CPU models; amount of memory; operating system; names and versions of relevant software libraries and frameworks. (yes/partial/no)
\item This paper formally describes evaluation metrics used and explains the motivation for choosing these metrics. (yes/partial/no)
\item This paper states the number of algorithm runs used to compute each reported result. (yes/no)
\item Analysis of experiments goes beyond single-dimensional summaries of performance (e.g., average; median) to include measures of variation, confidence, or other distributional information. (yes/no)
\item The significance of any improvement or decrease in performance is judged using appropriate statistical tests (e.g., Wilcoxon signed-rank). (yes/partial/no)
\item This paper lists all final (hyper-)parameters used for each model/algorithm in the paper’s experiments. (yes/partial/no/NA)
\item This paper states the number and range of values tried per (hyper-) parameter during development of the paper, along with the criterion used for selecting the final parameter setting. (yes/partial/no/NA)
\end{itemize}